\documentclass[letterpaper, 10 pt, conference]{ieeeconf}  

\IEEEoverridecommandlockouts                              

\usepackage{graphicx} 
\usepackage{times} 
\usepackage{amsmath} 
\usepackage{amssymb}  
\usepackage{multirow,color}
\usepackage{cite}
\usepackage{algpseudocode}
\usepackage{varwidth}
\usepackage{subfig}
\usepackage{booktabs}

\usepackage{color}

\renewcommand{\baselinestretch}{0.98}

\title{\Large \bf
A Handheld Device for the In Situ Acquisition of Multimodal Tactile Sensing Data
}

\author{Joshua Wade, Tapomayukh Bhattacharjee, and Charles C. Kemp
\thanks{J. Wade, T. Bhattacharjee, and C. C. Kemp are with the Healthcare Robotics Lab at Georgia Tech. Contact josh\_wade@gatech.edu .}}

\begin{document}

\maketitle
\thispagestyle{empty}
\pagestyle{empty}

\begin{abstract}
Multimodal tactile sensing could potentially enable robots to improve their performance at manipulation tasks by rapidly discriminating between task-relevant objects. Data-driven approaches to this tactile perception problem show promise, but there is a dearth of suitable training data. In this two-page paper, we present a portable handheld device for the efficient acquisition of multimodal tactile sensing data from objects in their natural settings, such as homes. The multimodal tactile sensor on the device integrates a fabric-based force sensor, a contact microphone, an accelerometer, temperature sensors, and a heating element. We briefly introduce our approach, describe the device, and demonstrate feasibility through an evaluation with a small data set that we captured by making contact with 7 task-relevant objects in a bathroom of a person's home. 
\end{abstract}





\section{Introduction}
During manipulation, a robot can potentially benefit by using tactile sensing to recognize the task-relevant object with which it has made contact. For example, a robot could attempt to reduce the force it applies to a particular type of object, such as a person’s body, or it could attempt to maneuver its end effector with respect to a target object that it is attempting to grasp, such as an object on top of a counter. Data-driven approaches for rapid tactile perception have shown promise \cite{BhattacharjeeKapustaRehgKemp2013}, but suitable training data is lacking.

\begin{figure}
\centering
\includegraphics[height=5.5cm]{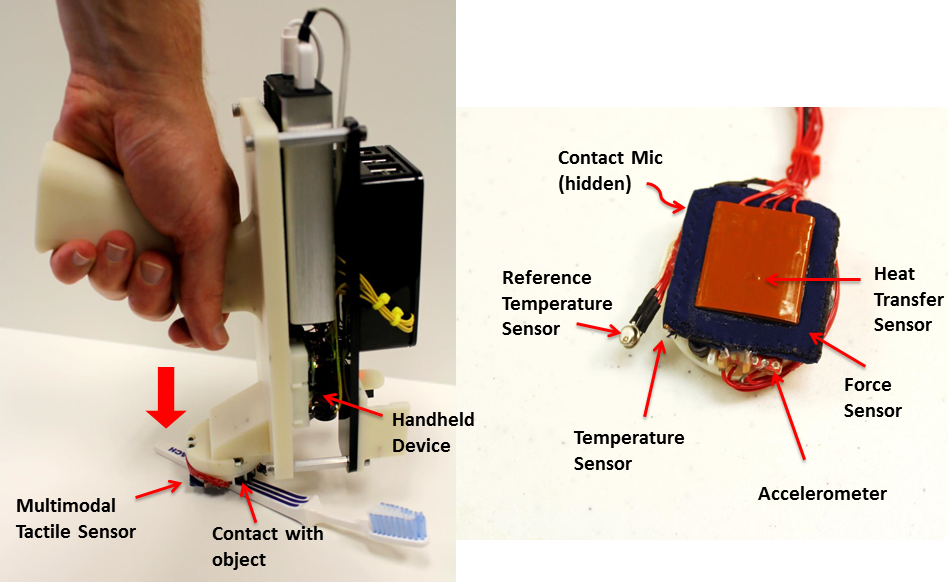}
\caption{\label{fig:device}\textit{Left: Handheld data acquisition device, Right: The multimodal tactile sensor that makes contact with objects.}}
\vspace{-0.3cm}
\end{figure}

To help address this challenge, we have developed a portable handheld device (see Figure \ref{fig:device}) for the efficient acquisition of multimodal tactile sensing data from objects in their natural settings. Robot vision and audition, including face detection and speech recognition, have benefited greatly from large labeled data sets of pictures, videos, and audio collected by people. Our motivation is to enable people to efficiently acquire tactile training data for robots, so that tactile perception systems for robots can similarly benefit. In this paper, we briefly describe the device and demonstrate the feasibility of our approach through an evaluation with a small data set captured from 7 task-relevant objects in a bathroom of a person’s home. Our evaluation focuses on distinguishing a target object, which we refer to as the tactile foreground, from a clutter object, which we refer to as the tactile background. Each foreground/background pair corresponds with two objects relevant to a specific task, such as placing a towel on a towel rack or picking up a toothbrush from a counter.

\begin{figure*}[t!]
\centering
\includegraphics[width=13.3cm]{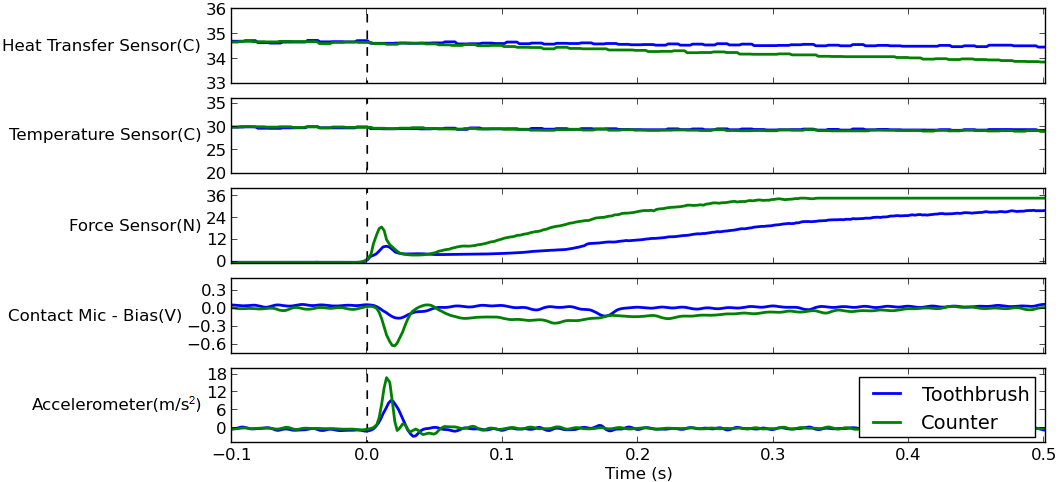}
\caption{\label{fig:graph}\textit{Multimodal tactile sensor response for contact with a toothbrush resting on a counter and the counter itself.}}
\vspace{-0.3cm}
\end{figure*}

\section{Description of the Device}\label{sec:setup}

Figure \ref{fig:device} (Left) shows the complete handheld data acquisition device. Figure \ref{fig:device} (Right) shows the multimodal tactile sensor that mounts to the front of the handheld device and comes into contact with objects. The multimodal tactile sensor includes a sensor for  measuring heat transfer, a fast response thermistor for temperature sensing, a fabric-based tactile sensor for force estimation, and a contact microphone and accelerometer mounted behind these elements to measure vibration and acceleration. We also mounted an LM35 Precision Centigrade Temperature Sensor that does not make contact with the object to measure the ambient air temperature and to serve as a reference when calibrating the fast response thermistor and the heat transfer sensor's thermistor. 

The handheld device uses an onboard camera to save a picture of each object for documentation. It also uses a Sharp digital IR proximity sensor and timer to estimate the average velocity over the last 6.5 cm of motion before contact. This serves to characterize the speed at which the user makes contact with an object.  The onboard Raspberry Pi 2 and 8 channel 12 bit ADS7828 analog-to-digital converter (ADC) record data to a USB flash drive from the force sensor, contact microphone, and accelerometer at 500 Hz and from the heat-transfer sensor and temperature sensor at 100 Hz. 



We based the sensor for measuring heat transfer on our work in \cite{bhattacharjeematerial}. The sensor uses a Thorlabs HT10K - Flexible Polyimide Foil Heater with a 10 kOhm Thermistor \cite{thorlabs}. Unlike our previous work, we also used an EPCOS fast response 10K NTC thermistor to measure the air temperature before contact and the object's temperature during contact. We converted the raw ADC output from the thermistor in the heat transfer sensor and the fast response thermistor to degrees Celsius using a third-order polynomial fit ($R^2=0.994$) based on calibration data with the LM35 Precision Centigrade Temperature Sensor. 



The force sensing modality uses a single 1 inch square taxel in a voltage divider circuit based on our stretchable fabric-based tactile sensor described in \cite{bhattacharjee2013tactile}. We converted the raw ADC output from the taxel to force in newtons, assuming a uniform pressure distribution over the taxel, using a third-order polynomial fit ($R^2=0.984$) with calibration data collected using an ATI Mini45 Force/Torque sensor. 

We used a 20 mm diameter piezo disc, also known as a contact microphone, in a voltage divider circuit with a 10 mOhm resistor to measure an object's acoustic response upon contact. We mounted the contact microphone in a compartment within the multimodal tactile sensor's 3D-printed base.

To measure acceleration as the sensor makes contact, we used an ADXL335 accelerometer with a sensing bandwidth set to 500 Hz. Note that due to the Nyquist rate, sampling at greater than 1000 Hz would be more appropriate for the accelerometer, but would exceed our device's capabilities.


\section{Evaluation}\label{sec:exp}
We selected common household objects found in a bathroom that are associated with activities of daily living (ADLs), which are tasks with which an assistive robot might provide beneficial help. Using our device, we captured data for the following task-relevant tactile foreground versus tactile background recognition problems: toothbrush vs. counter; towel vs. towel rack; toilet handle vs. toilet tank; and toilet seat vs. toilet tank. The first author used the handheld device to make contact with each of the 7 objects used in these 4 recognition problems 10 times for a total of 70 distinct trials. He allowed the heat-transfer sensor to reheat for 3 minutes before the first trial with an object and then waited for 10s between subsequent trials with the object. Each trial lasted approximately 5s. When making contact, he attempted to move the device in a linear motion normal to the surface of the object. He intentionally varied the speed at which he moved the device for each trial with an object in order to capture varying contact conditions. Figure \ref{fig:graph} shows 0.5s of the multimodal sensor data from a trial with the toothbrush and a trial with the counter. As seen in the figure, contact with the stiffer, effectively immobile counter produces a larger magnitude response with a steeper initial slope from the force sensor, contact microphone, and accelerometer, when compared with the signals resulting from contact with the more compliant and mobile toothbrush.


%
%
\section{Analysis and Results}\label{sec:analysis}
For this initial evaluation, we did not use the heat transfer and temperature data, although we plan to investigate their value in the future. We truncated the raw time series from the force sensor, contact microphone and acceleration modalities to include 2100 time samples from approximately 0.2s before contact to 4s after contact. We then normalized each modality by subtracting the mean and dividing by the variance across all of the modality's data, after which we vectorized each modality and concatenated the resulting vectors into a single vector. To reduce the effect of noise and overfitting, we computed a low-dimensional representation of the data using principal component analysis (PCA) with 15 principal components, which accounted for more than 97\% of the variance in the data. We used  a support vector machine (SVM) classifier with a linear kernel and 5-fold cross-validation to recognize each object pair. A summary of the results are shown in Table \ref{tbl:summary_table}. 


\begin{table}[t!]
\caption{Performance Summary.\label{tbl:summary_table}}
\begin{center}
\vspace{-0.4cm}
\begin{tabular} {|c|c|c|}
\hline
Foreground & Background & Recognition \\
   Object & Object & Accuracy \\
\hline
\hline
Toothbrush & Counter & 100\%\\
\hline
Towel & Towel Rack & 95\%\\
\hline
Toilet Handle & Toilet Tank & 75\%\\
\hline
Toilet Seat & Toilet Tank & 75\%\\
\hline

\end{tabular}
\end{center}
\vspace{-0.8cm}
\end{table}






\textit{\textbf{Acknowledgment}}: 
This work was supported in part by NSF Awards EFRI-1137229 and IIS-1150157, and benefited from discussions with Vincent Dureau as part of a Google Faculty Research Award.

%
%

\bibliographystyle{IEEEtran}

\bibliography{references}

\end{document}